\documentclass[conference]{IEEEtran}
\IEEEoverridecommandlockouts

\usepackage{cite}
\usepackage{graphicx}
\usepackage{psfrag} 
\usepackage{bm}
\usepackage{subfigure}
\usepackage{amsmath,amssymb,amsfonts}
\usepackage{epsfig}
\usepackage{amsmath}
\usepackage{multirow}
\usepackage{lineno}
\usepackage{algorithm}
\usepackage{algorithmic} 
\usepackage{diagbox} 
\usepackage[margin=0.7in]{geometry}
\usepackage{lipsum}

\usepackage{xcolor}

\newcommand{\p}[1]{\mathop{\mbox{\it p} } }

\renewcommand{\vec}[1]{\ensuremath{\boldsymbol{#1}}}
\newcommand{\be}{\begin{equation}}
\newcommand{\ee}{\end{equation}}
\newcommand{\ba}{\begin{array}}
\newcommand{\ea}{\end{array}}
\newcommand{\bea}{\begin{eqnarray}}
\newcommand{\eea}{\end{eqnarray}}
\newcommand{\bean}{\begin{eqnarray*}}
\newcommand{\eean}{\end{eqnarray*}}

\newcommand{\rmh}{^{\rm \dag}}
\newcommand{\rmt}{^{\rm T}}

\definecolor{white}{rgb}{1,1,1}

\title{Dual-Attention Based 3D Channel Estimation}

\author{Xiangzhao Qin and Sha Hu\\
Lund Research Center\\ Huawei Technologies Sweden AB, Sweden.\\ 
email: \{chrisqin, hu.sha\}@huawei.com
}

\begin{document}
\maketitle

\begin{abstract}
For multi-input and multi-output (MIMO) channels, the optimal channel estimation (CE) based on linear minimum mean square error (LMMSE) requires three-dimensional (3D) filtering. However, the complexity is often prohibitive due to large matrix dimensions. Suboptimal estimators approximate 3DCE by decomposing it into time, frequency, and spatial domains, while yields noticeable performance degradation under correlated MIMO channels. On the other hand, recent advances in deep learning (DL) can explore channel correlations in all domains via attention mechanisms. Building on this capability, we propose a dual attention mechanism based 3DCE network (3DCENet) that can achieve accurate estimates.   

\end{abstract}

\section{Introduction}

Artificial intelligence (AI) and deep learning (DL) are potential enablers for the sixth generation (6G) systems~\cite{QW24,ZQ24, Mehran2019Deep,Jiang2021Dual, Sun2020Learn, SC23,Raviv2023Modular, Zhou2021RCNet}. Emerging proposals consider using AI to replace or enhance those essential modules in communication systems such as channel estimation (CE)~\cite{MK21, LL17, SS19, LT23} and multi-input and multi-output (MIMO) detection~\cite{Schmid2022LowComplexity,Honkala2021DeepRx, Faycal2022E2E, Zhou2023Graph2023}. An AI receiver example is depicted in Fig.~1. In advanced end-to-end (E2E) designs, transmit blocks can also be replaced with AI modules.

In the context of CE, conventional linear minimum-mean-square-error (LMMSE) estimator~\cite{LL14, CL15} works well in practice. However, an optimal CE requires three-dimensional (3D) processings, whose complexity can be prohibitive. Suboptimal estimators decompose 3DCE into completely separate filters in time, frequency, and spatial domains (3$\times$1D), or a joint 2D filter in time-frequency domains and a 1D filter in spatial domain (2D+1D). These simplifications can result in performance losses under correlated MIMO channels. On the other hand, DL based CE including super-resolution convolutional neural-network (SRCNN)~\cite{Mehran2019Deep}, ChannelNet~\cite{SS19}, enhanced deep super-resolution (EDSR)~\cite{LL17}, Channelformer~\cite{LT23}, and others have been proposed to approach LMMSE based CE, but there are still noticeable losses compared to the optimal 3DCE.

In this paper, we firstly derive the optimal 3DCE based on demodulation reference signal (DMRS) in the fifth-generation new-radio (5G-NR) system, and analyze the optimal noise-power allocation to different domains. Followed by theoretical analysis, we propose a 3DCE network (3DCENet) that explores channel correlations among all domains by merging the least-square (LS) estimates from all antenna-ports, time and frequency resource elements (REs) as an input feature. The 3DCENet is constructed with dual attention mechanisms comprising of a spatial attention (SA) and a time-frequency attention (TFA) that enhances filter operations on spatial and time-frequency domains, respectively. Numerical results show that the proposed 3DCENet outperforms conventional genie 2D and (2D+1D) based CE up to 4dB in terms of mean-square-error (MSE), and is superior than SRCNN and EDSR based 2D estimators.

 \subsubsection*{Notations:} The operators $(\cdot)^{T}$, $(\cdot)^{\dag}$, and $\otimes$ denote the transpose, Hermitian conjugate, and Kronecker product, respectively. The operation $\text{vec}(\bm{H})$ vectorizes a matrix $\bm{H}$ by stacking the column vectors on top of each other, and $\text{vec}^{-1}(\cdot)$ is the inverse operation.

 \begin{figure}
            \hspace{2mm}
    \centering
    \includegraphics[width = 0.48 \textwidth, keepaspectratio]{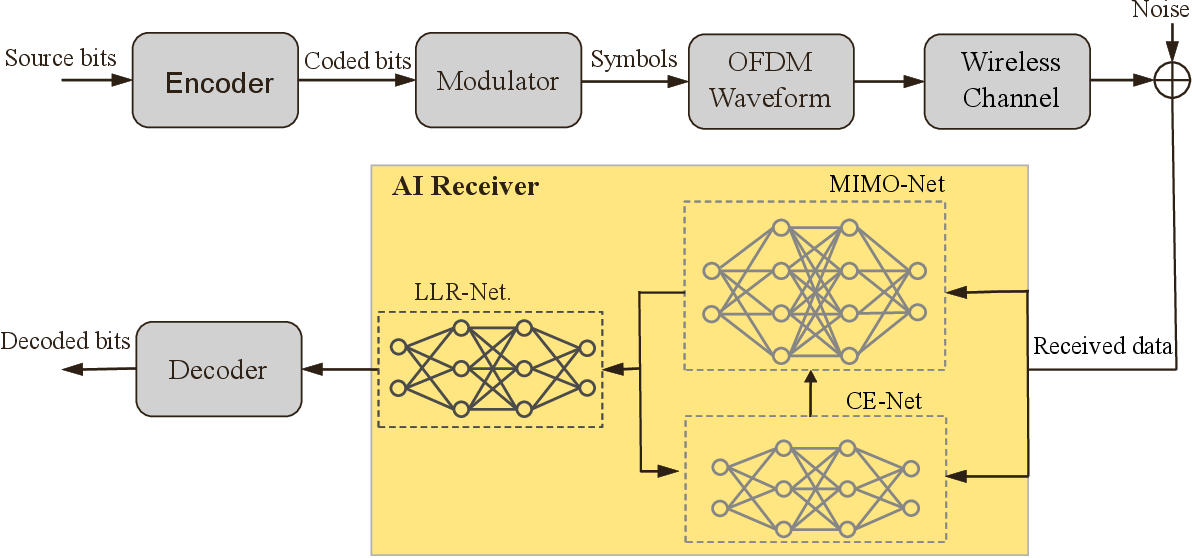}
        \vspace{-1mm}
    \caption{An AI receiver design comprised of AI based CE,  MIMO detection, and log-likelihood ratios (LLRs) computations~\cite{QW24, ZQ24}.}  
    \label{fig:AIRec}
    \vspace{-5mm}
\end{figure}

\section{CE for MIMO System}  

Consider an $N_{r} \!\times\! N_{t}$ MIMO system, where $N_{r}$ and $N_{t}$ denote the numbers of receive and transmit antennas, respectively. The transmit symbols and DMRS are assembled in OFDM symbols. Each subframe comprises of $N_{s}$ OFDM symbols, and each symbol consist of $N_{c}$ subcarriers (SCs). In total there are $N\!=\!N_{c}N_{s}$ resource elements (REs) in one subframe. Further, there are $N_{p}$ REs allocated with DMRS and $N_{d}$ REs transmits data, with $N_p\!+\!N_d\!=\!N$. A typical DMRS pattern is shown in Fig.~2. 

The received signal vector $\bm{y}_{m,n}(k)\! \in\! \mathbb{C}^{N_{p} \!\times\! 1}$ corresponds to the DMRS received on the $m$th antenna and transmitted from the $n$th antenna on the $k$th OFDM symbol, and $\bm{h}_{m,n}(k) \! \in \!\mathbb{C}^{N_{p} \times 1}$ is the channel vector to be estimated of the same size. Assuming $K$ symbols are used for CE,  the received signal and channel vectors are stacked together as
\bea  \bm{y}_{m,n}&\!\!\!=\!\!\!&[\bm{y}_{m,n}\rmt(1) ,\bm{y}_{m,n}\rmt(2),\cdots, \bm{y}_{m,n}\rmt(K)]\rmt,  \\
\bm{h}_{m,n}&\!\!\!=\!\!\!&[\bm{h}_{m,n}\rmt(1) ,\bm{h}_{m,n}\rmt(2),\cdots, \bm{h}_{m,n}\rmt(K)]\rmt. 
\eea
The received signal corresponding to DMRS reads
\begin{equation}\label{eqn:system_model_ce}
    \bm{y}_{m,n}= \bm{P}_{n}\bm{h}_{m,n}  + \bm{w}_{m,n} , 
\end{equation}  
where $\bm{P}_{n} \!\in\! \mathbb{C}^{KN_{p}\! \times\! KN_{p}}$ is a diagonal matrix whose diagonal elements carry DMRS from the $n$th transmit antenna. The noise $\bm{w}_{m,n}  \!\in \!\mathbb{C}^{KN_{p} \!\times\! 1}$ is modelled as additional white Gaussian noise (AWGN) with a zero-mean and a covariance matrix $\sigma_{w}^{2}\bm{I}$. The LS estimates of $\bm{h}_{m,n}$ is obtained as
\begin{equation}\label{eqn:h_ls}
    \hat{\bm{h}}_{m,n}^{\text{LS}} = \left(\bm{P}_{n}^{\dag}\bm{P}_{n}\right)^{-1}
    \bm{P}_{n}^{\dag}\bm{y}_{m,n}.
\end{equation}

\subsection{Correlation Models in Different Domains}
Further stacking the channel vectors $\bm{h}_{m,n}$ for different transmit and receiver antennas into a single vector $\bm{h}$ as
\bea \label{eqn:hvec}
    \bm{h}&\!\!\!\!\!=\!\!\!\!\!\!&\big[\bm{h}_{1,1}^{\rm T}, \bm{h}_{1,2}^{\rm T}, \cdots, \bm{h}_{1,N_t}^{\rm T}, \bm{h}_{2,1}^{\rm T}, \bm{h}_{2,2}^{\rm T}, \cdots, \bm{h}_{2,N_t}^{\rm T}, \cdots,  \notag \\
    &&\;\, \bm{h}_{N_r, 1}^{\rm T}, \bm{h}_{N_r,2}^{\rm T}, \cdots, \bm{h}_{N_r,N_t}^{\rm T} \big]\rmt.
\eea
The correlation matrix of $\bm{h}$ equals
\bea \label{R3D}
    \bm{R}_{\text{3D}}=\mathbb{E}\left\{ \bm{h} \bm{h}\rmh\right\}=\bm{R}_{s}\otimes\bm{R}_{tf}=(\bm{R}_{s, r}\otimes\bm{R}_{s, t})\otimes\bm{R}_{tf},
\eea
where $\bm{R}_{s}$ is the spatial correlation among antennas, and $\bm{R}_{tf}$ is the correlation matrix in time and frequency domains which is assumed to be identical for all transmit and receive antenna-pairs. The correlation matrix $\bm{R}_{s}$ is further decomposed into $\bm{R}_{s, r}$ and $\bm{R}_{s, t}$, which are correlations among receive and transmit antennas, respectively, and both can be modelled (with `$\ast$' denotes `$t$' or `$r$') as
\bea \label{eqn:Rtx}
    \bm{R}_{s, \ast}=\begin{bmatrix}
    r_{11}       & r_{12} & \cdots & \cdots & r_{1N_{\ast}} \\
    r_{21}       & r_{22} & \cdots & \cdots & r_{2N_{\ast}} \\
   \vdots       & \vdots  & \ddots & \ddots &  \vdots   \\
    r_{N_{\ast}1}       & r_{N_{\ast}2} & \cdots & \cdots & r_{N_{\ast}N_{\ast}}
\end{bmatrix}\!.
\eea 

On the other hand, the 2D correlation matrix on time and frequency domains reads
\bea \label{eqn:Rtf}
    \bm{R}_{tf}=\mathbb{E}\left\{ \bm{h}_{m,n} \bm{h}_{m,n}\rmh\right\}=\bm{R}_{t}\otimes\bm{R}_{f},
\eea
where $\bm{R}_{t}$ denotes channel correlation in time domain which is determined by the Doppler spread, and $\bm{R}_{f}$ denotes channel correlation in frequency domain which is determined by the frequency selectivity or power delay profile (pdp).

In 5G-NR system, the spatial correlation is modelled as an exponential model specified by a parameter $\alpha\!<\!1$ such that
\bea \label{spatcorr} r_{ij}=\alpha^{|i-j|} .\eea 
With Jakes model, for a time difference $\Delta t$, the correlation is
\bea  \bm{R}_{t}(\Delta t)=J_0(2\pi f_d \Delta t), \eea
where $J_0$ is the zeroth-order Bessel function of the first kind, and $f_d$ is the maximum Doppler shift. 

The frequency correlation is the discrete Fourier transform (DFT) of pdp, and with a frequency difference $\Delta f$, it equals
\bea  \label{freqcorr}  \bm{R}_{f}(\Delta f)=\sum_{\ell}\sigma_\ell^2 \exp(-2\pi \tau_\ell \Delta f), \eea
where $\tau_\ell$ and $\sigma_\ell^2$ are the delay and power of the $\ell$th channel path in time domain, respectively.

\subsection{The Optimal 3DCE}

A 3DCE exploits correlations across all domains in spatial, time, and frequency, described in (\ref{spatcorr})-(\ref{freqcorr}). With (\ref{R3D}), the optimal LMMSE based 3DCE of $\bm{h}$ reads
\bea \label{hest} \bm{h}_{\text{3D}}=\bm{W}_{\text{3D}} \hat{\bm{h}}^{\text{LS}} ,\eea 
where $\hat{\bm{h}}^{\text{LS}}$ is the LS estimate of $\bm{h}$. The filter equals
\bea \label{W3D} \bm{W}_{\text{3D}}=  \bm{R}_{\text{3D}}\mathcal{A}\rmt(\mathcal{A}\bm{R}_{\text{3D}}\mathcal{A}\rmt+\sigma_w^2\bm{I})^{-1}, \eea
where the selection matrix $\mathcal{A}$ is of sizes $KN_PN_rN_t\!\times\!NN_rN_t$. Denote a selection matrix $\mathcal{A}_{n}$  for the $n$th transmit antenna which is of sizes $KN_p\!\times\!KN$ and each row contains a single one where a DMRS is located and remaining entries are zeros. Letting $\text{Diag}\big[  \mathcal{A}_{1},   \mathcal{A}_{2}, \cdots, \mathcal{A}_{N_t}\big]$ denote a block-diagonal matrix whose diagonal blocks are $ \mathcal{A}_{m}$,  it holds that
\bea \label{An} \mathcal{A}= \bm{I}_{N_r}\!\otimes \text{Diag}\big[  \mathcal{A}_{1},   \mathcal{A}_{2}, \cdots, \mathcal{A}_{N_t}\big].\eea
 For the pattern in Fig.~2, $\mathcal{A}_{n}$ for the four transmit antennas are permuted versions of each other. It can be seen that if the optimal 3DCE is applied, a matrix inversion of sizes $KN_pN_rN_t\!\times\!KN_pN_rN_t$ for $\mathcal{A}\bm{R}_{\text{3D}}\mathcal{A}\rmt\!+\!\sigma_w^2\bm{I}$ is needed, whose complexity can be prohibitive.

\begin{figure}
    \centering
    \includegraphics[width = 0.32 \textwidth, keepaspectratio]{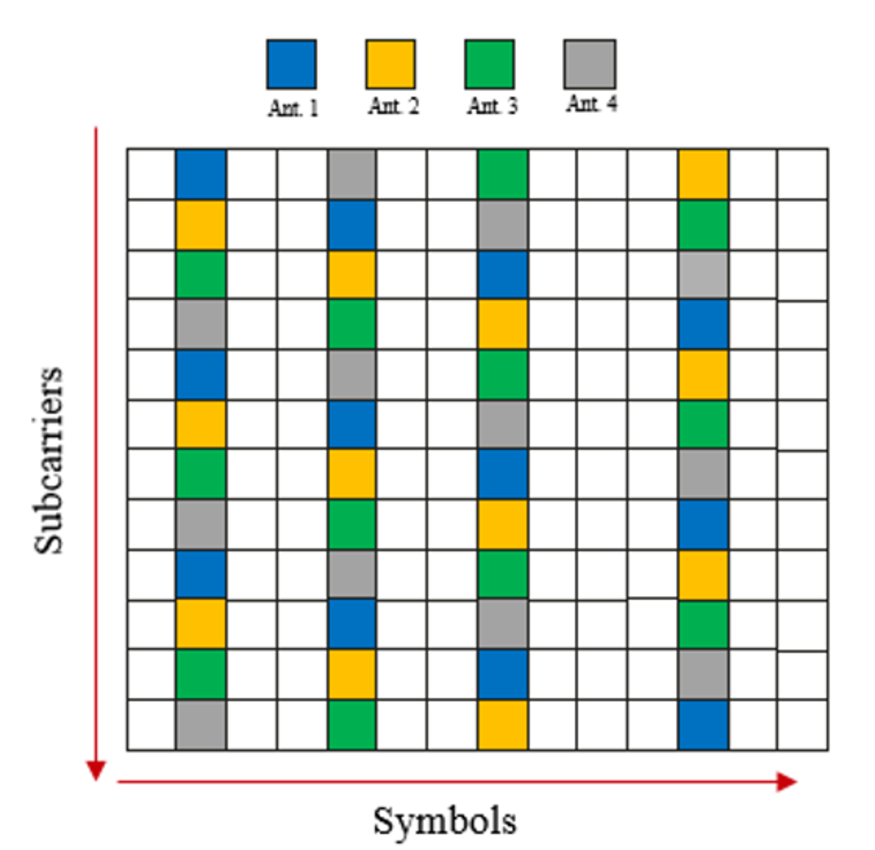}
        \vspace{-3mm}
    \caption{DMRS configurations for $N_{t}\!=\!4$, $N_f\!=\!12$, $N_s\!=\!14$, and  $N_p\!=\!3$.}  
    \label{fig:AIRec}
    \vspace{-5mm}
\end{figure}

\begin{figure*}
    \centering
    \includegraphics[width = 1 \textwidth, keepaspectratio]{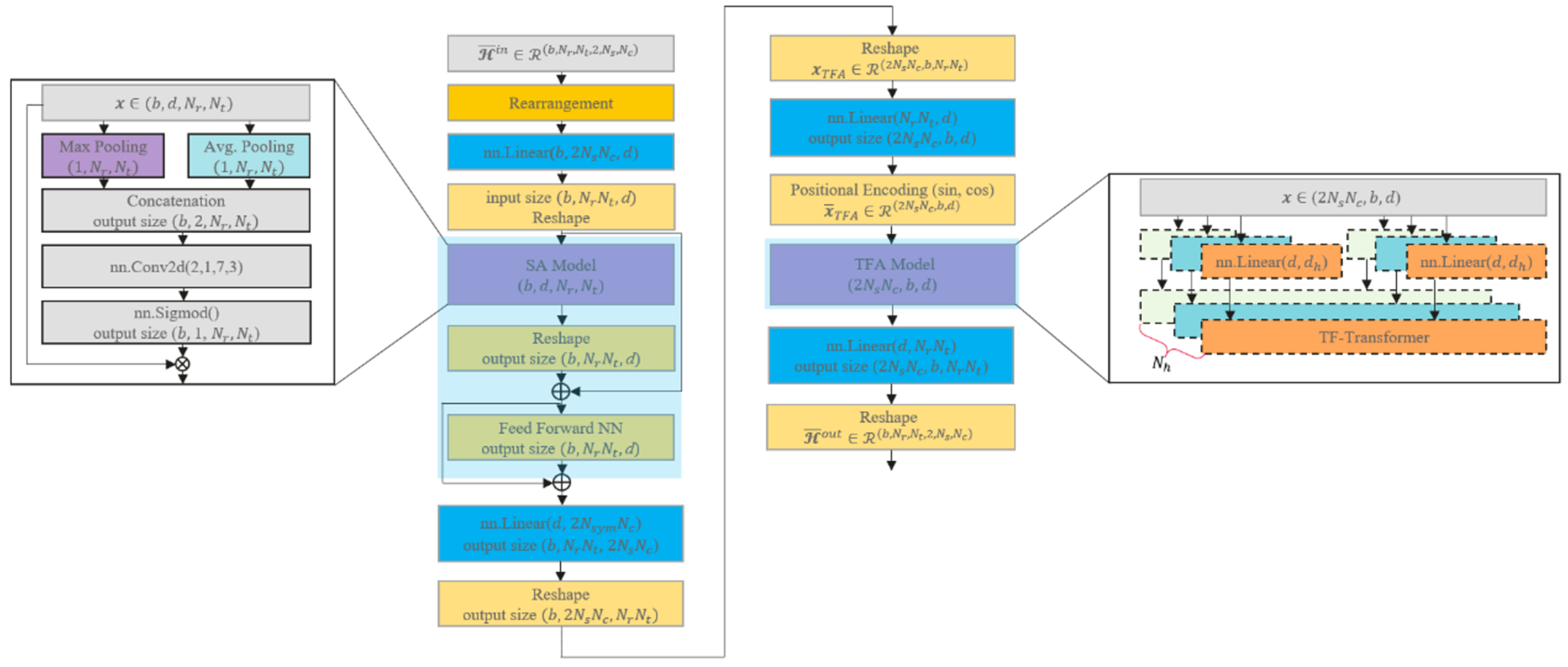}
        \vspace{-5mm}
    \caption{The proposed 3DCE design that comprises a two-stage process of SA followed by TFA modules. }  
    \label{fig:AIRec}
    \vspace{-5mm}
\end{figure*} 

\subsection{Suboptimal (2D+1D)-CE and (3$\times$1D)-CE }

A standard approach for simplifying 3DCE is to apply a joint 2D filtering on time and frequency domains, and a 1D filtering on spatial domain. Assuming $\mathcal{A}_{n}\!=\!\mathcal{A}_{1}$ for all $n$ in (\ref{An}), it holds that
\bea \label{An1} \mathcal{A}= \bm{I}_{N_rN_t}\otimes\mathcal{A}_1.\eea
Note that
\bea 
&&\!\!\!\!\!\!\!\!\!\!\!\!\!\!\!\! \!\!\!\!\!\! \bm{R}_{\text{3D}}\mathcal{A}\rmt\!=\! (\bm{R}_{s}\otimes\bm{R}_{tf})(\bm{I}_{N_rN_t}\otimes\mathcal{A}_1\rmt \!=\!\bm{R}_{s}\otimes(\bm{R}_{tf}\mathcal{A}_1\rmt),    \\
&&\!\!\!\!\!\!\!\!\!\!\!\!\!\!\!\!\!\!\! \!\!\!  \mathcal{A}\bm{R}_{\text{3D}}\mathcal{A}\rmt\!=\!\bm{R}_{s}\otimes( \mathcal{A}_1\bm{R}_{tf}\mathcal{A}_1\rmt). 
\eea 
The inversion can be approximated as
\be \label{Dapprox}  (\mathcal{A}\bm{R}_{\text{3D}}\mathcal{A}\rmt\!+\!\sigma_w^2\bm{I})^{-1}\!\approx\! (\bm{R}_{s}+\sigma_{s}^2\bm{I})^{-1}\otimes(\mathcal{A}_1\bm{R}_{tf}\mathcal{A}_1\rmt+\sigma_{tf}^2\bm{I})^{-1},
\ee
and the 3DCE filter in (\ref{W3D}) is
\bea \label{W3D1} \hat{\bm{W}}_{\text{3D}}\approx \bm{W}_s\otimes \bm{W}_{tf} , \eea
where
\bea 
 \bm{W}_s\!\!\!&=&\!\!\!\bm{R}_{s}(\bm{R}_{s}+\sigma_{s}^2\bm{I})^{-1},\\
  \bm{W}_{tf}\!\!\!&=&\!\!\!\bm{R}_{tf}\mathcal{A}_1\rmt(\mathcal{A}_1\bm{R}_{tf}\mathcal{A}_1\rmt+\sigma_{tf}^2\bm{I})^{-1}.
\eea
The parameters $\sigma_{s}^2$ and $\sigma_{tf}^2$ denote the noise powers in spatial and time-frequency domains, respectively. Inserting (\ref{W3D1}) back into (\ref{hest}) yields
\bea \label{hest2}
\bm{h}_{\text{3D}}\approx ( \bm{W}_s\otimes \bm{W}_{tf}) \hat{\bm{h}}^{\text{LS}}=\text{vec}(\bm{W}_{tf}\hat{\bm{H}}_{\text{3D}}\bm{W}_s\rmt) ,\eea 
where
\bea \hat{\bm{H}}_{\text{3D}}=\text{vec}^{-1}({\hat{\bm{h}}^{\text{LS}} }). \eea

It can be readily seen that the 3DCE can be approximated by a 2D filtering in time-frequency domain and a 1D filtering on spatial domain. Note that the sub-optimality of (2D+1D)-CE lies in the fact that the tensor decomposition (\ref{Dapprox}) is not exact and renders accuracy losses.

Note that the inversion of $\mathcal{A}_1\bm{R}_{tf}\mathcal{A}_1\rmt+\sigma_{tf}^2\bm{I}$ when computing the 2DCE is still of sizes $KN_p\!\times\!KN_p$. To further reduce complexity and following the same analysis, it can be further decomposed into two 1D filters on time and frequency domains separately. To be specific, the 2DCE for the $n$th transmit antenna, denoted as $\bm{W}_{tf,n}$, is approximated as
\bea \label{Wtfn} \hat{\bm{W}}_{tf,n}\approx \bm{W}_{t,n}\otimes \bm{W}_{f,n} , \eea
where
\bea 
 \bm{W}_{t,n}\!\!\!&=&\!\!\!\bm{R}_{t}\mathcal{A}_{nt}\rmt(\mathcal{A}_{nt}\bm{R}_{t}\mathcal{A}_{nt}\rmt+\sigma_{t}^2\bm{I})^{-1},\\
  \bm{W}_{f,n}\!\!\!&=&\!\!\!\bm{R}_{f}\mathcal{A}_{nf}\rmt(\mathcal{A}_{nf}\bm{R}_{f}\mathcal{A}_{nf}\rmt+\sigma_{f}^2\bm{I})^{-1}.
\eea
The variables $\sigma_{t}^2$, $\sigma_{f}^2$ denote the noise powers on time and frequency domains, respectively. The matrices $\mathcal{A}_{nt}$ of sizes $K_p\!\times\!K$  and  $\mathcal{A}_{nf}$ of sizes $N_p\!\times\!N_c$ are selection matrices for the $n$th transmit antenna on time and frequency domains, respectively, and each row contains a single one where a DMRS symbol is located, and the remaining entries are zeros. 

Different simplifications of 2DCE and comparisons to ($2\!\times\!1$D)-CE have been well studied in for OFDM systems and there are many existing literatures, see e.g.~\cite{CL15}. In general, there are some slight degradations from the decomposition (\ref{Wtfn}), but it can be well justified by the complexity reductions. Hence, we will mainly focus on comparisons between the optimal 3DCE and (2D+1D)-CE, and examine the potential gains by exploring the spatial domain with DL based designs.

\subsection{Optimal Noise Power Allocation}

A motivation to consider designing the AI based 3DCE is due to the noise- power splitting issue with conventional approaches, which is typically overlooked. As shown in (\ref{Dapprox}), the noise power $\sigma_w^2$ is split into the spatial and the time-frequency domains as $\sigma_s^2$ and $\sigma_{tf}^2$, respectively. It is suboptimal to set them both to $\sigma_w^2$ , which enhances the noise significantly. Our observation is that $\sigma_s^2$ and $\sigma_{tf}^2$ should be jointly designed, and one heuristic principle can be to minimize the trace difference between $\mathcal{A}\bm{R}_{\text{3D}}\mathcal{A}\rmt\!+\!\sigma_w^2\bm{I}$ and $(\bm{R}_{s}\!+\!\sigma_{s}^2\bm{I})\otimes(\mathcal{A}_1\bm{R}_{tf}\mathcal{A}_1\rmt\!+\!\sigma_{tf}^2\bm{I})$. 

Note that
\bea 
&&\!\!\! \!\!\!\!\!\!\!\!\! (\bm{R}_{s}+\sigma_{s}^2\bm{I})\otimes(\mathcal{A}_1\bm{R}_{tf}\mathcal{A}_1\rmt+\sigma_{tf}^2\bm{I})-  \mathcal{A}\bm{R}_{\text{3D}}\mathcal{A}\rmt   \notag \\&&\!\!\! \!\!\! \!\!\!  =\sigma_{s}^2(\bm{I}\otimes(\mathcal{A}_1\bm{R}_{tf}\mathcal{A}_1\rmt +\sigma_{tf}^2\bm{I}))+\sigma_{tf}^2((\bm{R}_{s}+\sigma_{s}^2\bm{I})\otimes \bm{I}) \notag \\ &&  \!\!\! \!\!\! \!\!\! +\sigma_{s}^2\sigma_{tf}^2\bm{I}.
\eea 
To minimize trace difference, it should hold that
\bea \label{noisepowereq}
\sigma_{w}^2=\sigma_{s}^2+\sigma_{tf}^2+\sigma_{s}^2\sigma_{tf}^2.
\eea 
This shows that both $\sigma_{s}^2$ and $\sigma_{tf}^2$ are smaller than $\sigma_{w}^2$. If setting $\sigma_{s}^2\!=\!\sigma_{tf}^2\!=\!\sigma_{\ast}^2$, the optimal solution is
\bea  \label{optnoisepower}
\sigma_{\ast}^2=\sqrt{\sigma_{w}^2+1}-1.
\eea

Similar issue exists when further decomposing (\ref{Wtfn}), and the noise power $\sigma_{tf}^2$ allocated in time and frequency domains is split into $\sigma_{t}^2$ and $\sigma_{f}^2$, respectively, which should also be optimized similarly.

\section{3DCENet Based on Dual Attention} 
 
Dual attention mechanism~\cite{BY23} learns multi-domain features via customized attention mechanisms, where spatial correlation is processed by a spatial attention (SA) and correlations from delay-Doppler are handled by a time-frequency attention (TFA). Due to the sub-optimality of (2D+1D)-CE, we apply dual attention mechanism and propose the 3DCENet that comprises of SA and TFA, with a detailed paradigm in Fig.~3. 

\subsection{Preprocess of LS Estimates}

The first step is a preprocess to interpolate LS estimate on DMRS to data REs. The input tensor $\bm{\mathcal{H}}^{p}\!\in\!\mathbb{C}^{N_{p} \times N_{r} \times N_{t}}$ is constructed as
\begin{equation}\label{eqn:H_p}
    \bm{\mathcal{H}}^{p}{\left[:,m,n\right]} = \hat{\bm{h}}_{m,n}^{\text{LS}}. 
\end{equation}
A Gaussian tensor $\bm{\Pi} \!\in \!\mathbb{C}^{N_{d} \!\times\! N_{p}\times N_{r} \times N_{t}}$ interpolates $\bm{\mathcal{H}}^{p}$ on the first dimension to $\bm{\mathcal{H}}^{\text{in}} \!\in\! \mathcal{C}^{N_{d} \times N_{r} \times N_{t}}$ as 
\begin{equation}\label{eqn:H_in}
    \bm{\mathcal{H}}^{\text{in}} = \bm{\Pi} \cdot \bm{\mathcal{H}}^{p}.
\end{equation} 
Then, $ \bm{\mathcal{H}}^{\text{in}}$ is passed via an SRCNN represented by $\mathcal{F}_{\mathrm{SRCNN}}\left(\cdot\right)$, and the output tensor $\bm{\mathcal{H}}^{\text{out}} \! \in  \!\mathcal{C}^{N_{d} \times N_{r} \times N_{t}}$ reads
\begin{equation}\label{eqn:H_out}
    \bm{\mathcal{H}}^{\text{out}} = \mathcal{F}_{{\mathrm{SRCNN}}}\left(\bm{\mathcal{H}}^{\text{in}}\right)\!.
\end{equation}

\subsection{Spatial Attention (SA)}

After preprocess and combining the estimates $\bm{\mathcal{H}}^{\text{out}}$ on data REs and $\bm{\mathcal{H}}^{p}$ on DMRS, a complete channel tensor $ \bm{\mathcal{H}}^{\text{out}}$ is formed as a 6D tensor with dimensions $(b, N_r , N_t , 2  , N_s  , N_c )$, where $b$ denotes the batch size and `2' represent the real and imaginary parts. Then, time and frequency domains together with real and imaginary parts are merged together such that it becomes a 4D tensor $\bar{\vec{H}}$ with dimensions $(b, N_r , N_t , 2N_s N_c )$. Then, $\bar{\vec{H}}$ is compressed in the last dimension to $\bar{\vec{x}}$ with dimensions $(b, N_r , N_t , d)$, and $\bar{\vec{x}}$ is permuted to $\vec{x}$ with dimensions $(b, d, N_r , N_t )$  and sent to SA module.

In SA module, $\vec{x}$ is further compressed via a max-pooling layer and an average-pooling layer to obtain $\vec{x}_{\text{mean}}$ and $\vec{x}_{\text{max}}$, respectively, both with dimensions $(b, 1, N_r , N_t) $ and concatenated as $\underline{\vec{x}}\!=\![\vec{x}_{\text{mean}}, \vec{x}_{\text{max}}]$. Then, $\underline{\vec{x}}$ is passed via a convolution-layer to obtain a scaling tensor with dimensions $(b, 1, N_r , N_t )$, which is multiplexed to the input $\vec{x}$. The output from the SA module has the same dimensions as $\vec{x}$, and multiple SA models can be concatenated sequentially. After the SA process, the obtained tensor is reshaped and passed through a feed-forward network (FFN) and a linear-layer and recovered into a 4D tensor $\vec{x}_{\text{TFA}}$ with dimensions $(2N_s N_c , b,  N_r , N_t )$, as shown in Fig.~3.

\subsection{Time-Frequency Attention (TFA)}

Before TFA module, $\vec{x}_{\text{TFA}}$ is further compressed in spatial domain ($N_r$ and $N_t$), followed by a positional encoding (PE) to obtain a tensor $\bar{\vec{x}}_{\text{TFA}}$ with dimensions $(2N_s N_c , b,  d)$. Later, $\bar{\vec{x}}_{\text{TFA}}$ is processed with multiple transformer encoder blocks to explore correlations in time and frequency domains.

Since the complete 3DCENet design is jointly trained, it enhances conventional (2D+1D)-CE design by optimizing the interactions between SRCNN2D, SA, and TFA modules to approach the optimal 3DCE. The training is based on labels $\bm{\mathcal{H}}^{\text{label}}\! \in \!\mathbb{C}^{N_{d} \times N_{r} \times N_{t}}$ and MSE losses. After 3DCENet, the estimates of $\bm{\mathcal{H}}$ is sent to the MIMO detection module as shown in Fig.~1.

\subsection{Complexity of SA and TFA Modules}

The computational complexity of 3DCENet is dominated by the SA and TFA modules.  Below we summarize the number of trainable parameters and the multiply–accumulate 
operations (MACs) per batch for each module.

\paragraph{SA}
The SA block consists of three components: (i) a channel compression layer mapping $2N_{s}N_{c}$ input channels to $d$ channels; (ii) a fixed convolution layer $\mathrm{Conv2D}(2,1,7,3)$ used to generate spatial attention weights; and (iii) a two-layer FFN with hidden dimension $d_{\mathrm{ff}}$.  
The convolution layer contains exactly $2 \!\times\! 1\!\times\! 7^{2}\! +\! 1 \!=\! 99$ trainable parameters and requires $98N_{r} N_{t}$ MACs per batch. 

\paragraph{TFA.}
The TFA block is a standard transformer encoder operating on a sequence of length $L \!=\! 2N_{s}N_{c}$. Each encoder blocks includes multi-head self-attention (MHA) and a two-layer FFN. The MHA contributes the quadratic term $2L^{2} d$ from correlation and weighted sum, and $4Ld^{2}$ from projections. The FFN has a complexity of $2Ld d_{\mathrm{ff}}$, which dominates the overall complexity.  

Combining all components, the total number of weights and MACs are shown in Table~\ref{tab:SA_TFA_complexity}.

\begin{table}[t]
\caption{Number of Parameter and MAC operations of the 3DCENet.}
\label{tab:SA_TFA_complexity}
\centering
\renewcommand{\arraystretch}{1.2}
\begin{tabular}{|c|c|c|}
\hline
\textbf{Module} &
\textbf{Number of Weights} &
\textbf{MACs per Batch} \\
\hline

SA  &
\(
\begin{aligned}
& (2N_sN_c)d +2dd_{\mathrm{ff}} \\
&+ d_{\mathrm{ff}} +2d+99
\end{aligned}
\)
&
\(
\begin{aligned}
& 2N_rN_t (dN_sN_c + d d_{\mathrm{ff}}+49)
\end{aligned}
\)
\\
\hline

TFA  &
\(
\begin{aligned}
& 4d^2 + 5d+ 2d d_{\mathrm{ff}}+ d_{\mathrm{ff}}
\end{aligned}
\)
&
\(
\begin{aligned}
&
2L(2 d^2 + L d + dd_{\mathrm{ff}})
\end{aligned}
\)
\\
\hline

\end{tabular}
  \vspace{-4mm}
\end{table}

\section{Numerical Simulation}  

The 3DCENet is constructed as in Fig.~3 with $d\!=\!512$ in SA module. In TFA module, 4 transformer encoders are applied with 4 hidden layers each, and the model size and FFN size are both set to 512. This yields about 6M trainable parameters and 3G Flops per CE. The MIMO channel is modelled as extended typical urban (ETU) and extended pedestrian-A (EPA) channels~\cite{3GPPChannelModel} in time domain, and with a medium spatial correlation specified by $\alpha\!=\!0.3$ as in (\ref{spatcorr}) for both transmit and receive antennas .

In Fig.~4 and 5, MSEs are evaluated under correlated $4\!\times\!4$ MIMO and ETU100Hz channel. The 3DCE is approximated by (2D+1D)-CE as in (\ref{W3D1}). As seen, (2D+1D)-CE with $\sigma_{tf}\!=\!\sigma_{s}\!=\!\sigma$ outperforms the genie 2DCE about 1dB. Noticeably, there is additional 1.4dB gains with the optimal power allocation $\sigma^{\ast}$ in (\ref{optnoisepower}). Further, it is also better than setting $\sigma_{tf}\!=\!\sigma$ and a small $\!\sigma_{s}\!=\!0.1\sigma^2$. In Fig.~5, 3DCENet significantly outperforms conventional AI based SRCNN2D and EDSR2D estimators. Moreover, (2D+1D)-CE with optimal noise power shows 3dB gains over the genie 2DCE, and 3DCENet provides another 1dB enhancements.

In Fig.~6, BLERs are shown under correlated $2\!\times\!2$ MIMO and EVA70Hz channel, with 16QAM modulation. As seen, 3DCENet together with a graph neural network aided expectation propagation (GEPNet)~\cite{Zhou2023Graph2023} detector performs close to 2DCE with optimal maximum likelihood (ML) detector. It is also 2dB better than a conventional 2DCE with EP detector, and outperforms SRCNN2D-CE with GEPNet by 0.7dB.

\begin{figure}[t]
    \centering
            \hspace*{-2mm}
    \includegraphics[width = 0.43 \textwidth, keepaspectratio]{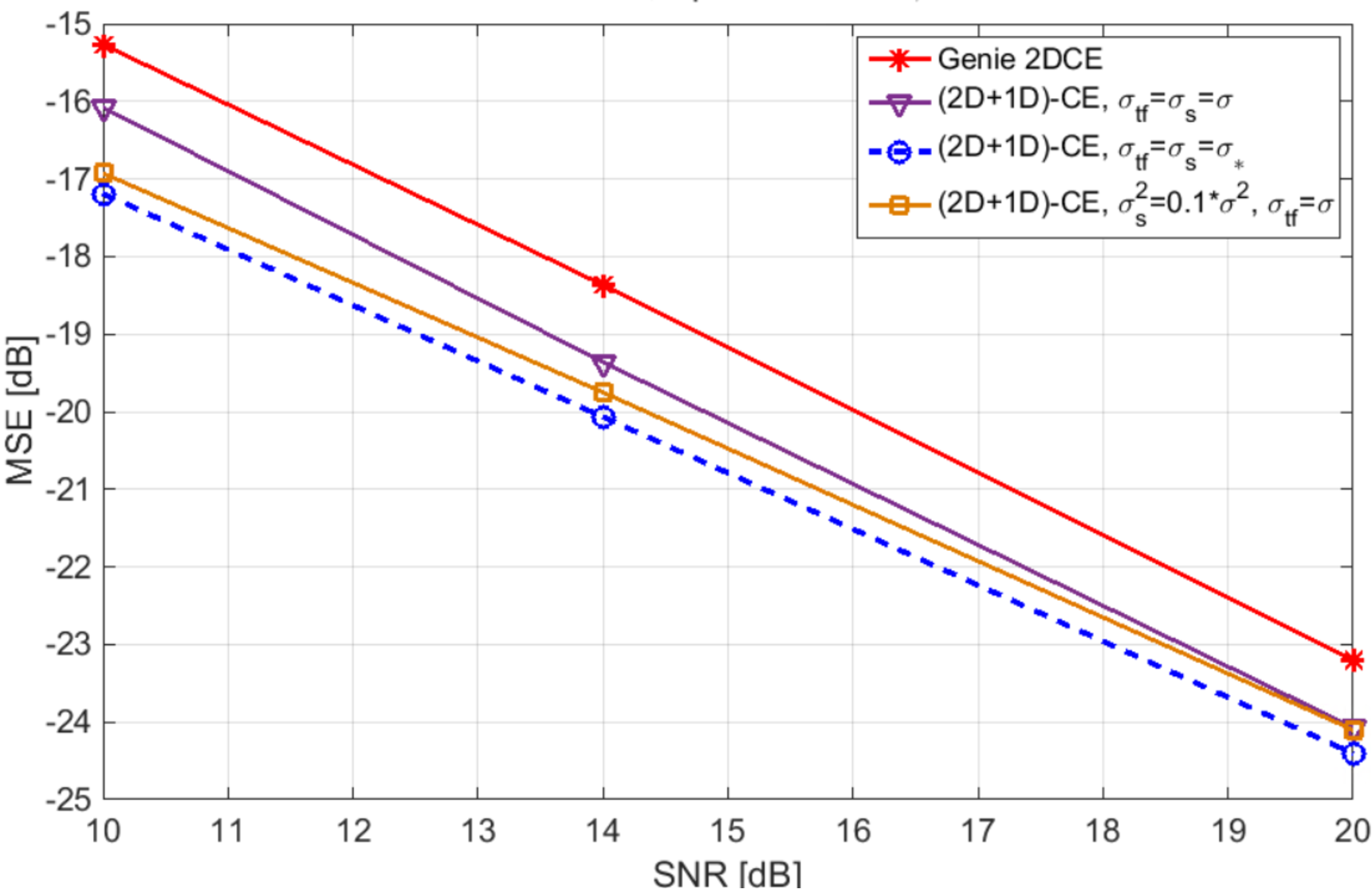}
        \vspace{-3mm}
    \caption{CE-MSEs with different noise power allocations under correlated $4\!\times\!4$ MIMO and ETU100Hz channel. }  
    \label{fig:AIRec}
    \vspace{-2mm}
\end{figure} 

\begin{figure}
    \centering
                \hspace*{-2mm}
    \includegraphics[width = 0.44 \textwidth, keepaspectratio]{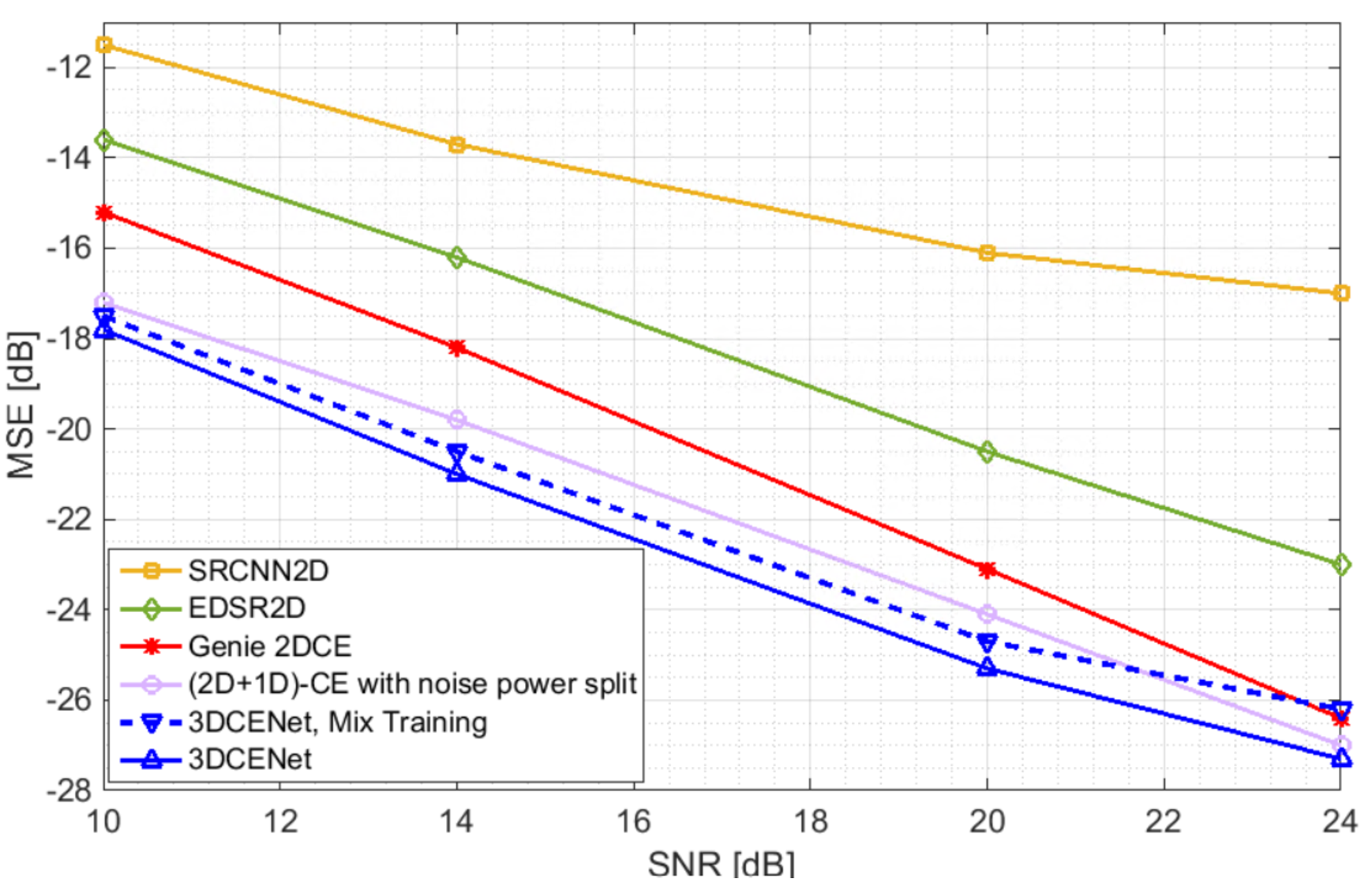}
        \vspace{-3mm}
    \caption{CE-MSEs with the proposed 3DCENet and other methods. }  
    \label{fig:AIRec}
    \vspace{-5mm}
\end{figure} 

\begin{figure}[t]
    \centering
                \hspace*{4mm}
    \includegraphics[width = 0.44 \textwidth, keepaspectratio]{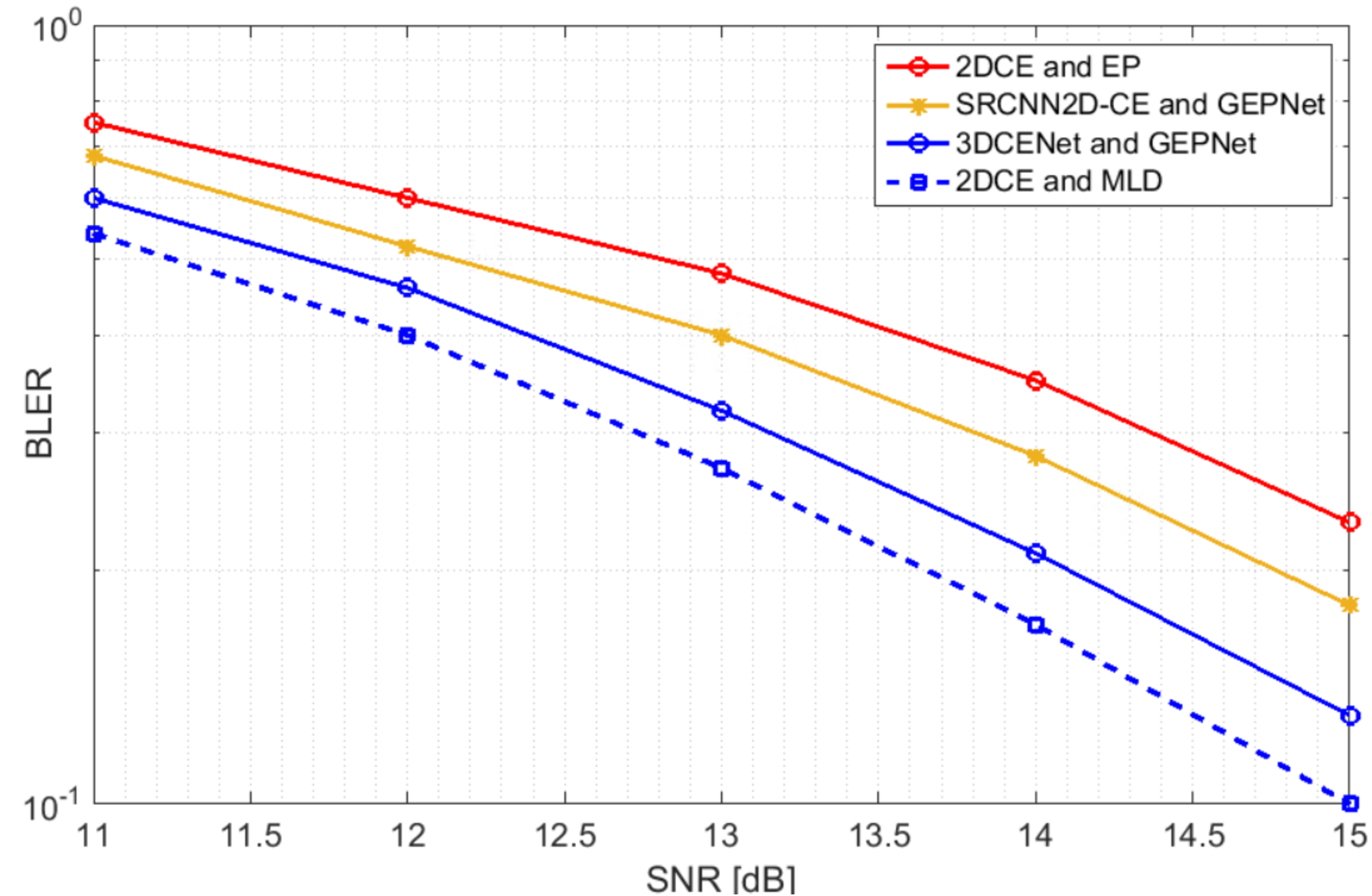}
        \vspace{-3mm}
    \caption{BLER of $2\!\times\!2$ MIMO and 16QAM modulation. }  
    \label{fig:AIRec}
    \vspace{-5mm}
\end{figure}

\section{Conclusions}  

We have derived the optimal noise power-allocation in different domains for conventional 3DCE, and proposed a double attention mechanism based 3DCENet to enhance (2D+1D)-CE in MIMO systems . Numerical results have shown that  3DCENet can achieve accurate CE and outperforms both conventional and AI based 2DCE designs under correlated MIMO channel by fully exploiting correlations in all domains.

\bibliographystyle{IEEEtran}

\end{document}